\pdfoutput=1

\documentclass[11pt]{article}

\usepackage[]{acl}

\usepackage{times}
\usepackage{latexsym}
\usepackage{bbm}
\usepackage{multirow}
\usepackage{amsmath}
\usepackage{graphicx}
\usepackage{booktabs}
\usepackage{amssymb}

\usepackage[T1]{fontenc}

\usepackage[utf8]{inputenc}

\usepackage{microtype}

\usepackage{inconsolata}

\title{Insert or Attach: Taxonomy Completion via Box Embedding}

\author{
Wei Xue$^{1}$, Yongliang Shen$^{1\dagger}$, Wenqi Ren$^{2}$, Jietian Guo$^{2}$, Shiliang Pu$^{2}$, Weiming Lu$^{1\dagger}$\\
$^{1}$College of Computer Science and Technology, Zhejiang University \\
$^{2}$Hikvision Research Institute\\
\texttt{\{lokilanka, syl, luwm\}@zju.edu.cn} \\
\texttt{\{renwenqi, guojietian, pushiliang.hri\}@hikvision.com}
}

\begin{document}
\maketitle

\renewcommand{\thefootnote}{\fnsymbol{footnote}}
\footnotetext[2]{\;Corresponding author.}
\renewcommand{\thefootnote}{\arabic{footnote}}

\begin{abstract}
Taxonomy completion, enriching existing taxonomies by inserting new concepts as parents or attaching them as children, has gained significant interest. Previous approaches embed concepts as vectors in Euclidean space, which makes it difficult to model asymmetric relations in taxonomy. In addition, they introduce pseudo-leaves to convert attachment cases into insertion cases, leading to an incorrect bias in network learning dominated by numerous pseudo-leaves. Addressing these, our framework, \textsc{TaxBox}, leverages box containment and center closeness to design two specialized geometric scorers within the box embedding space. These scorers are tailored for insertion and attachment operations and can effectively capture intrinsic relationships between concepts by optimizing on a granular box constraint loss. We employ a dynamic ranking loss mechanism to balance the scores from these scorers, allowing adaptive adjustments of insertion and attachment scores. Experiments on four real-world datasets show that \textsc{TaxBox} significantly outperforms previous methods, yielding substantial improvements over prior methods in real-world datasets, with average performance boosts of 6.7\%, 34.9\%, and 51.4\% in MRR, Hit@1, and Prec@1, respectively.
\end{abstract}

\section{Introduction}
Taxonomy, a critical knowledge graph with an "is-a" relationship, plays a vital role in information retrieval, recommendation systems, and question answering~\cite{chatterjee2022analysing, chuang2003automatic, kejriwal2022local, Kerschberg2001tax, suchanek2007yago, huang2019taxo, yang2017ea, yu-etal-2021-technical}.
However, manual taxonomy enrichment is inefficient and costly due to the constant emergence of new concepts. To address the challenge of incorporating new concepts, taxonomy completion has been introduced, with new concepts either inserted as both parents and children or attached only as children~\cite{jiang2022taxoenrich, zhang2021taxonomy, wang2022qen, zeng2021taxogen}. This task goes beyond taxonomy expansion, which primarily treats new concepts as leaf nodes and tends to have limitations in downstream applications~\cite{shen2020taxoexpan, liu-etal-2021-temp, yu2020steam, manzoor2020expanding, phukon2022team, boxtaxo}. 
\begin{figure}[]
    \centering
    \includegraphics[width=\linewidth]{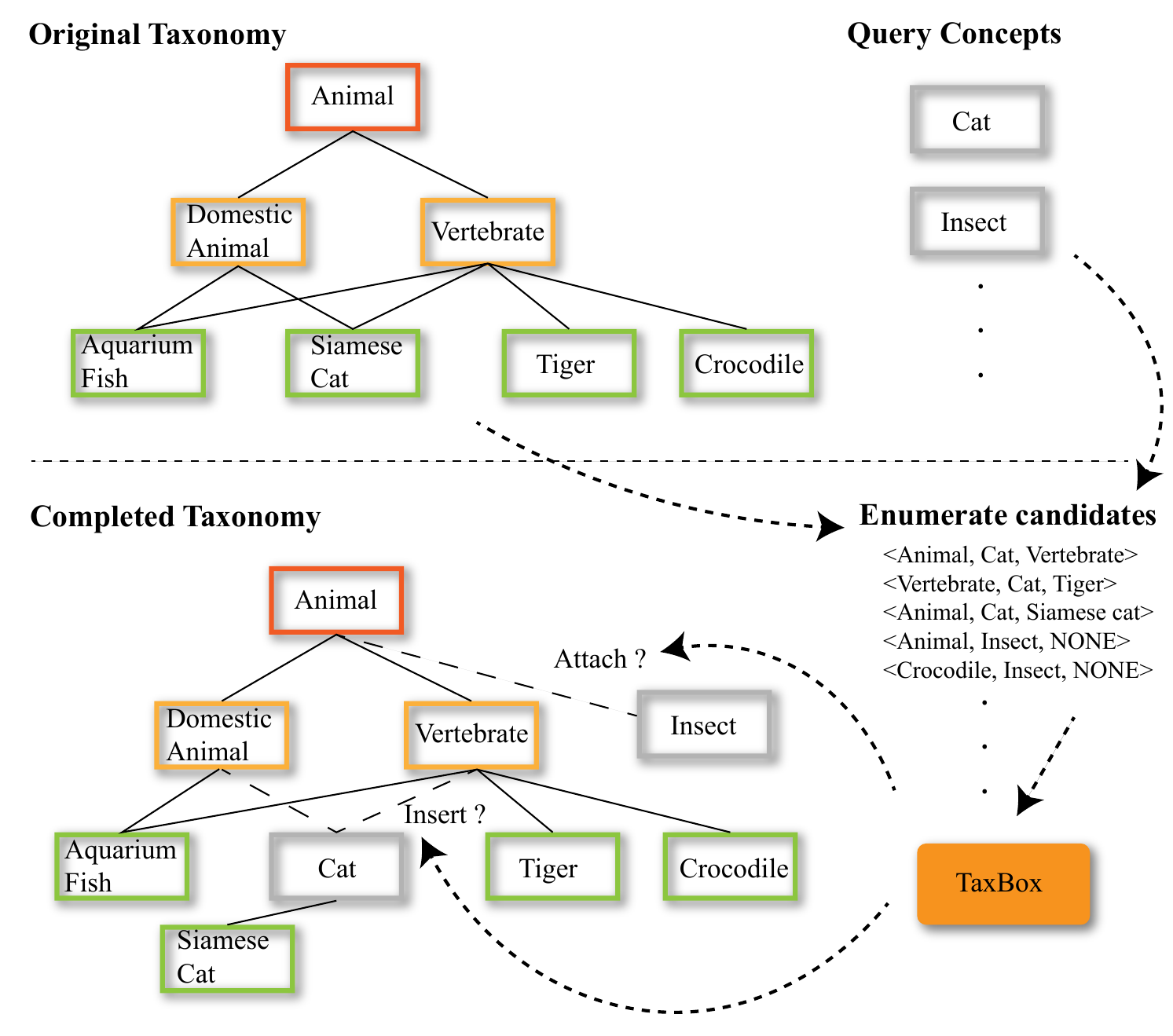}
    \caption{Example of taxonomy completion with our \textsc{TaxBox} framework.}
    \label{fig:fig1}
\end{figure}

Taxonomy completion entails a more comprehensive incorporation of new concepts with two operations: insertion and attachment. For instance, in Figure \ref{fig:fig1}, new query concepts such as \textit{cat} and \textit{insect} are added to the existing \textit{animal} taxonomy. The process requires enumerating all possible candidate positions within the original taxonomy, including existing edges like \textit{<Animal, Vertebrate>} and implicit edges from each node to its descendants such as \textit{<Animal, Tiger>}. Each candidate position is then paired with the query concept, and a confidence score is calculated. Finally, \textit{insect} is attached as a child of \textit{animal} and \textit{cat} is inserted as a parent of \textit{Siamese cat} and children of \textit{Domestic Animal} and \textit{Vertebrate} according to their confidences.

Recent research on taxonomy enrichment has examined various practical methods ~\cite{jiang2022taxoenrich, zhang2021taxonomy, wang2022qen, zeng2021taxogen}. Nevertheless, all of these approaches embed concepts as vectors in Euclidean space, which makes them less capable of modeling the asymmetric relationship ("is-a") in taxonomy. BoxTAXO \cite{boxtaxo} tried to employ box embedding, a representation method that can capture more prosperous and asymmetric relationships like inclusion, disjoint, and proximity among concepts through its geometric properties. However, this method is limited in real-world applications for its reliance only on the volume property, rendering it suitable only for the taxonomy expansion and even incapable of discerning optimal ancestor concepts and handling multiple parents during inference. Moreover, methods for taxonomy completion \cite{zhang2021taxonomy, wang2022qen} suffer from using a "pseudo-leaf" as a child node in attachment cases, leading to confusion in the matching. It is attributed that attachment cases often predominate due to leaf nodes’ prevalence in real taxonomies. Therefore, learning too much about the pseudo-leaf in the attachment cases may reduce the network's perceptual ability for child nodes in the insertion cases.

To overcome these limitations, we present a novel framework for taxonomy completion called \textbf{\textsc{TaxBox}}, which is the first to apply box embedding to taxonomy completion. This approach adopts a structurally enhanced box decoder, representing concepts as box embeddings~\cite{vilnis2018probabilistic} encompassing the information of children, furnishing richer semantics. Most importantly, \textsc{TaxBox} combines two probabilistic scorers to unify the process of insertion and attachment in the box embedding space and incorporates both the volume and center closeness properties of box embedding. Such a design effectively exploits the fine-grained geometric attributes of box embeddings, circumventing the need for a pseudo-leaf and yielding optimal, feasible results during the ranking process. Additionally, we propose two novel training objectives, optimizing both box volume and position, and rectifying scorer numerical imbalances. 

The specific contributions of this paper are outlined as follows:

\begin{itemize}
\item We introduce \textsc{TaxBox}, the first framework using box embedding for taxonomy completion with a structurally enhanced box decoder.
\item We establish insertion and attachment scorers, obviating the need for pseudo-leaves and ensuring the determination of optimal results.
\item We design box constraint loss, focusing on both volume and center closeness, and dynamic ranking loss, rectifying scorer numerical imbalance.
\item Experimental outcomes from four datasets demonstrate our model's efficacy, achieving 6.7\% MRR, 34.9\% Hit@1, and 51.4\% Prec@1 improvements over the previous methods.
\end{itemize}

\begin{figure*}[htbp]
  \centering
  \includegraphics[width=\linewidth]{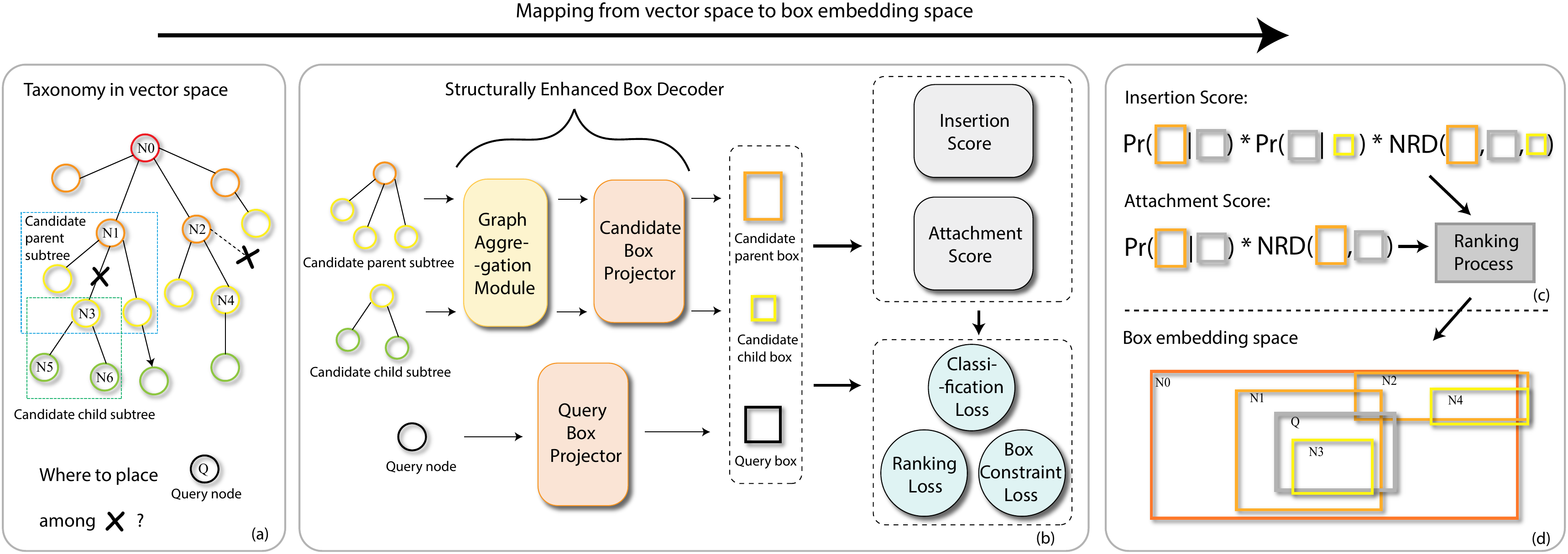}
  \captionof{figure}{Overview of \textsc{TaxBox} architecture. (a) The seed taxonomy tree with a query concept. (b) A structurally enhanced box decoder maps concepts among all the candidates and the query concept to the box embedding space. (c) Two probabilistic scorers calculate confidence of insertion or attachment for each candidate position. (d) Find the best position via ranking to complete the seed taxonomy with the novel concept in box embedding space.}
  \label{fig:fig2}
\end{figure*}

\section{Related work}
\textbf{Taxonomy Expansion and Completion.} Taxonomy expansion, the process of attaching  novel concepts into an existing taxonomy, has evolved over time with various approaches~\cite{shen2018hiexpan, shen2020taxoexpan, yu2020steam, manzoor2020expanding, liu-etal-2021-temp, ma2021hyperexpan, phukon2022team, boxtaxo}. Although effective, these methods have limitations in addressing real-world applications. Thus, \citet{zhang2021taxonomy} introduced taxonomy completion, a generalization that allows for the insertion of a concept as a parent to existing nodes, generating wider-reaching solutions. Subsequent research~\cite{wang2022qen, jiang2022taxoenrich, zeng2021taxogen} sought to tackle this more challenging version of taxonomy expansion. \citet{jiang2022taxoenrich} incorporated contextual embeddings into input embeddings, leveraging dual LSTMs to encode ancestor and descendant information~\cite{staudemeyer2019lstm}. Meanwhile, \citet{zeng2021taxogen} devised a generative strategy that concurrently generates concept names and classifies valid candidate positions. \citet{wang2022qen} introduced the Quadruple Evaluation Network (QEN), which utilized pretrained language models (PLM)~\cite{devlin2018bert, sanh2019distilbert} to augment initial embeddings with semantically rich term representations. \citet{arous2023taxocomplete} learns a position-enhanced node representation through anchor sets to better find the candidate.

\noindent \textbf{Box Embedding.} Box embedding represents a mapping technique that embeds concepts or objects within hyperplane boxes. Initially proposed by \citet{vilnis2018probabilistic}, this approach employs probabilistic box lattices to encapsulate entities in knowledge graphs as \textit{n}-dimensional rectangles. Subsequently, various studies have applied box embedding across diverse domains. For instance, \citet{rau2020predicting} predicted visual overlap in images, while \citet{onoe2021modeling} and \citet{patel2021modeling} focused on entity typing and multi-label classification, respectively. Moreover, \citet{dasgupta2022word2box} mapped words to capture set-theoretic semantics, and \citet{hwang2022event} and \citet{messner2022temporal} explored relation extraction and knowledge graph completion. These works highlight box embedding’s suitability for nuanced semantic relationship modeling.

\section{Preliminary}
Box embedding~\cite{vilnis2018probabilistic, chheda2021box} refers to a mapping that represents a concept or object as a hyperplane box.
A box $x=[x_m, x_M]$ is a hyperrectangle such that $x_m\in \mathbb{R}^d$ and $x_M\in \mathbb{R}^d$ where $x_m$ and $x_M$ represent the minimum and maximum endpoints of the box respectively along the $d$ axis and $x_{m,i} \leq x_{M,i}$ holds for each axis $i \in \{1,2,...,d\}$. The center of box embedding is formulated as:
\begin{align}
    \label{eq:eq0}
    \text{Cen}(x)=\frac{x_M+x_m}{2}
\end{align}

There are two important operations: \textit{Intersection} and \textit{Volume} which are required for the calculation of the conditional probability of boxes' containment. Given two box embedding $x = [x_m, x_M]$, $y=[y_m, y_M]$, the \textit{Intersection} of them is defined as follows:
\begin{align}
    \label{eq:eq1}
    \text{Inter}(x, y) = \left[\text{max}(x_m, y_m), \text{min}(x_M, y_M)\right]
\end{align}
where $min(\cdot , \cdot)$ and $max(\cdot, \cdot)$ in Equation \ref{eq:eq1} perform element-wise operations. Specifically, $min(a,b)= [min(a_1,b_1),..., min(a_d, b_d)]$, and similarly for $max(\cdot , \cdot)$. The \textit{Volume} is defined as:
\begin{align}
    \begin{split}
        \label{eq:eq2}
        \text{Vol}(x) = \prod_{i=1}^d \tau * &\text{softplus}(\frac{x_{M_i}-x_{m_i}}{\tau})\\
        \text{softplus}(a)&=\text{log}(1+\text{exp}\ a)
    \end{split}
\end{align}
where $\tau$ is a hyperparameter to adjust the smoothness.
The probability of box $x$ containing box $y$ or the conditional probability of $x$ given $y$ is:
\begin{align}
    \label{eq:eq3}
    \text{Pr}(x | y)&=\frac{\text{Vol}(\text{Inter}(x,y))}{\text{Vol}(y)}
\end{align}

\section{The \textsc{TaxBox} Framework}
In this section, we elaborate on the proposed \textsc{TaxBox} framework, as shown in Figure \ref{fig:fig2}. We begin by defining the problem in Section \ref{sec4.1}. Then, in Section \ref{sec4.2}, we introduce the structurally enhanced box decoder, which maps concepts into box embeddings with hierarchical information enhanced. Section \ref{sec4.3} focuses on the discussion of two probabilistic scorers that evaluate the query and candidate boxes, providing attachment and insertion scores. Finally, in Section \ref{sec4.4}, we elucidate the learning objectives that contribute to improved optimization of box decoding and scorer balancing.

\subsection{Problem Definition}
\label{sec4.1}
A taxonomy is a directed acyclic graph and is defined as $\mathcal{T}^0=(\mathcal{N}^0,\mathcal{E}^0)$ where each node $n \in \mathcal{N}^0$ represents a concept and each edge $\langle p,c \rangle \in \mathcal{E}^0$ represents the "is-a" relationship edge between concepts. Given a seed taxonomy $\mathcal{T}^0$ and a set of new concepts $C$, the definition of taxonomy completion is to construct a new taxonomy $\mathcal{T}=(\mathcal{N}, \mathcal{E})$ where $\mathcal{N}=\mathcal{N}^0\cup \mathcal{C}$ and $\mathcal{E}$ is updated by adding new edges among $\mathcal{C}$ and $\mathcal{N}^0$. To fulfill the task, all the candidate positions $\mathcal{P}=\{\langle p,c \rangle|\forall p \in \mathcal{N}^0, \forall c \in descendants(p)\}$ have to be evaluated given a novel concept $n\in \mathcal{C}$. The whole training paradigm follows self-supervised learning. For each node in the seed taxonomy, we pretend it to be a query and optimize it with a reconstructed taxonomy without the node.


\subsection{Structurally Enhanced Box Decoder}
\label{sec4.2}
The structurally enhanced box decoder includes a graph aggregation module to aggregate the hierarchical features from the ego subtree, as well as two box projectors map aggregated features and query embedding to box embedding space, respectively. An ego subtree of node $n$ is defined as a tree only containing $n$ and its one-hop children, denoted by $\mathbb{T}(n)$.
\begin{figure}[htbp]
    \centering
    \includegraphics[width=\linewidth]{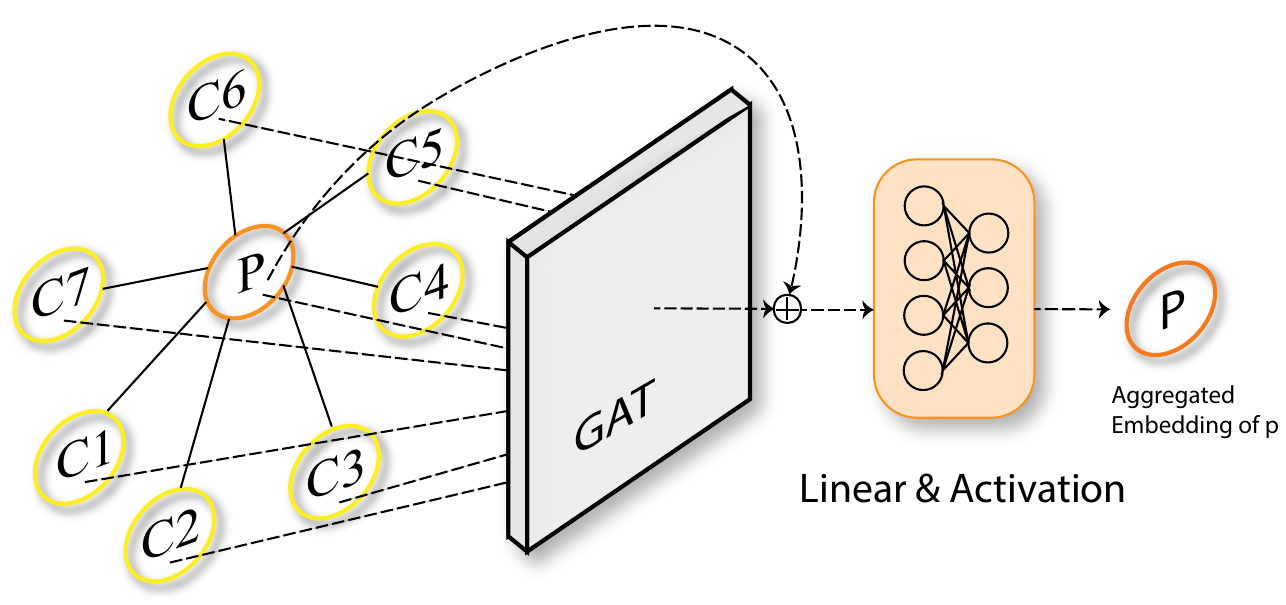}
    \caption{Details of Graph aggregation module.}
    \label{fig:figg}
\end{figure}

For a query $q$ and a possible candidate $\langle p,\ c \rangle \in \mathcal{P}$, we first obtain the embedding of each concept in the candidate along with their hierarchical information. As illustrated in Figure \ref{fig:figg}, we design a graph aggregation module to achieve this. The formulation is given by Equation \ref{eq:eq4}:
\begin{align}
    \begin{split}
        \label{eq:eq4}
    F_k=&\text{Lin}(\mathbb{R}(\text{GAT}(\mathbb{T}(k)) + \mathbb{T}(k))), k \in \{p, c\}
    \end{split}
\end{align}
where $F_k$ is the aggregated feature and $\mathbb{R}(\cdot)$ is a readout method, which implies that we only read out the root embedding of an ego subtree. \textit{Lin} denotes a linear layer with activation. To effectively fuse more information from relevant child nodes, we opt for \textit{GAT} (Graph Attention Network)~\cite{gat} to aggregate these trees in our implementation. 

Next, two box projectors with identical Highway network\cite{srivastava2015highway} structure project aggregated features and query embedding to box embeddings, respectively, as formulated in Equation \ref{eq:eq5}. To avoid potential conflicts arising from different latent spaces, we do not use a shared weight module for the aggregated parent/child features and query embedding.
\begin{align}
    \begin{split}
        \label{eq:eq5}
        B_q &= \text{QProjector}(F_q) \\
        B_k &= \text{CProjector}(F_k),\ k\in \{p, c\}
    \end{split}
\end{align}
where $F_q$ denotes query embedding and $B_q, B_p, B_c$ represent the box embedding of query, candidate parent, and candidate child, respectively. QProjector is the query box projector, and CProjector is the candidate box projector.

\subsection{Insertion and Attachment Scorer}
\label{sec4.3}
To make the best use of the geometric properties of box embedding like volume and center closeness, we design insertion scorer and attachment scorer to separately give confidence corresponding to these two cases.

\noindent \textbf{Insertion Scorer}. Assumes that our model captures fine-grained semantic relationships between two concept boxes optimized by box constraint loss (Section \ref{sec4.4}). Given a query concept $n$, we first introduce its positive candidate set $C_{pos}(n)=\{\langle p,\ c \rangle|\forall p \in \mathbb{P}(n), \forall c \in \mathbb{C}(n)\}$ and negative candidate set $C_{neg}(n)=\{\langle p,\ c \rangle| \exists p \notin \mathbb{P}(n)\ 	\lor \ \exists c \notin \mathbb{C}(n)\}$ where $\mathbb{P}$ and $\mathbb{C}$ refers to the parents and children of a node. Note that $\mathbb{C}(n)$ can be an empty set. For a positive candidate pair, the parent box can reliably hold the child box, while two boxes within a negative pair are disjoint. The closer the pair is in position, the more overlapping their box embedding will be. Based on this, we propose an insertion scorer ($S_I$) that represents the likelihood of performing insertion into the candidate as follows:
\begin{align}
    \begin{split}
        \label{eq:eq6}
    S_I(B_q, B_p, B_c) = &\text{Pr}(B_p|B_q) \cdot \text{Pr}(B_q|B_c)\\
    &\cdot \text{NRD}_{t}(B_q, B_p, B_c)
    \end{split}
\end{align}
where $\text{NRD}_{t}(\cdot,\cdot,\cdot)$ is the normalized reciprocal distance measuring the center closeness between the candidate parent and a query as well as that between the query and the candidate child. It is formulated as:
\begin{align}
    \begin{split}
        \label{eq:eq7}
        \text{RD}(B_q, B_{p_i}) &= \frac{1}{||\text{Cen}(B_q)-\text{Cen}(B_{p_i})||_2} \\
        \text{NRD}_{p}(B_q, B_p) &= \text{softmax}_{i=1}^n(\text{RD}(B_q, B_{p_i})) \\
        \text{NRD}_{t}(B_q, B_p, B_c) &= \text{NRD}_{p}(B_q, B_p)\\
        &\ \ \ \cdot \text{NRD}_{p}(B_q, B_c)
    \end{split}
\end{align}
where $\text{softmax}_{i=1}^n$ represents applying $softmax$ along a mini-batch and $B_{p_i}$ is a candidate in the mini-batch. $\text{RD}(\cdot, \cdot)$ is the reciprocal distance, and $\text{NRD}_p(\cdot, \cdot)$ only measures the closeness between the query and one side in the candidate.

\noindent \textbf{Attachment Scorer}. Similar to the insertion scorer, when faced with the scenario of a candidate pair with no child, an attachment scorer ($S_A$) is proposed. The attachment scorer is calculated as follows:
\begin{align}
    \begin{split}
        \label{eq:eq8}
        S_A(B_q, B_p)&=\text{Pr}(B_q|B_p)\cdot \text{NRD}_p(B_q,B_p)
    \end{split}
\end{align}


\subsection{Multiple Learning Objectives}
\label{sec4.4}
\noindent \textbf{Classification Loss}. The primary objective of our model is to determine the most suitable positions among all the candidate positions. We consider each candidate position as an independent category. Therefore, the problem can be reduced to a multi-label classification problem with a binary cross-entropy loss as:
\begin{align}
    \begin{split}
        \label{eq:eq9}
        \mathcal{L}_c &= -\frac{1}{|\mathcal{B}|}\sum_{(X_i,y_i)\in \mathcal{B}} y_i\text{log}(S_k(X_i))\\
        &+ (1-y_i)\text{log}(1-S_k(X_i)), k\in \{I, A\}
    \end{split}
\end{align}
where $X_i=(B_{q_i},B_{p_i},B_{c_i})$, $\mathcal{B}$ refers to a mini-batch consisting of one positive sample and several negative samples, $y\in \{0,1\}$ denotes whether the sample is positive or not. $S_k (k\in \{I, A\})$ means applying the insertion scorer if the candidate pair has both sides or the attachment scorer if it only has the parent side.

\noindent \textbf{Box Constraint Loss}. 
To better model the granularity of the "is-a" relationships amongst concepts using box embeddings, we focus on the geometric constraints originating from three properties of boxes: inclusion ($l_{in}$) and disjointness ($l_{dis}$) model the unidirectional relationships between two boxes, and centrality similarity ($l_{cen}$) facilitates scorers by obliging unrelated box pairs to assume orthogonal positions. Based on this, the loss functions for concept inclusion $L_{in}$ and disjoint $L_{dis}$ are as follows:
\begin{align}
    \begin{split}
        \label{eq:eq11}
        l_{in}(a,b)&= -\text{log}\ \text{Pr}(b|a)\\
        l_{dis}(a,b)&= \text{max}(0, \text{log}(1-\gamma(a,b)) \\
        &\ \ \ \ \ - \text{log}(1-\text{Pr}(a|b)))\\
        l_{cen}(a,b)&=  \text{max}(0, \text{log}(1-\gamma(a,b)) \\
        &\ \ \ \ \ - \text{log}(1-\text{Cen}(a)\cdot \text{Cen}(b)))\\
        L_{in}(a,b) &=  l_{in}(a,b) + l_{dis}(a,b)\\
        L_{dis}(a,b) &=  l_{dis}(a,b) + l_{dis}(b,a) + l_{cen}(a,b)
    \end{split}
\end{align}
The dynamic margin, $\gamma(a,b)$, between two concepts $a$ and $b$, is adapted from the Wu\&P similarity\cite{wu1994verb} and modulates their semantic distance:
\begin{align}
    \label{eq:eq10}
    \gamma(a,b)=\alpha \cdot \frac{2\times \text{depth}(\text{LCA}(a,b))}{\text{depth}(a)+\text{depth}(b)}
\end{align}
where $LCA(\cdot, \cdot)$ is the least common ancestor, $depth(\cdot)$ indicates the depth in the seed taxonomy, and $\alpha$ is a relaxation factor.
By imposing constraints on volume($l_{in}$, $l_{dis}$), position ($l_{cen}$), and distance ($S_{I/A}$), the optimization search space is effectively reduced.


Given a query box $B_q$, for a box $B_k(k\in \{p,c\})$ in a candidate, there are three possible scenarios: 1) $B_q$ is contained within $B_k$. 2) $B_k$ is contained within $B_q$. 3) both boxes are disjoint. When considering both sides of the candidate with a total of 6 possible cases, the box constraint loss is:
\begin{align}
    \begin{split}
        \label{eq:eq12}
        \mathcal{L}_b=&\frac{1}{|\mathcal{B}|}\sum_{(X_i,l_i)\in \mathcal{B}} l_{1i}\cdot L_{in}(B_{q_i}, B_{p_i})\\
        &+ l_{2i}\cdot L_{in}(B_{c_i}, B_{q_i})\\
        &+ l_{3i}\cdot L_{in}(B_{p_i}, B_{q_i})\\
        &+ l_{4i}\cdot L_{in}(B_{q_i}, B_{c_i})\\
        &+ (1-l_{1i})(1-l_{3i})\cdot L_{dis}(B_{q_i}, B_{p_i})\\
        &+ (1-l_{2i})(1-l_{4i})\cdot L_{dis}(B_{c_i}, B_{q_i})
    \end{split}    
\end{align}
where $l_i=(l_{1_i},l_{2_i},l_{3_i},l_{4_i})$ denotes whether the two sides of the candidate pair indeed contain the query concept or are contained by the query.

\begin{table*}[h]
    \centering
    \small
    \begin{tabular}{c|cccccccc}
        \toprule
         \multirow{2}{*}{\textbf{Method}} & \multicolumn{8}{c}{\textbf{MAG-CS}} \\
                \cline{2-9}
                & \textbf{MR $\downarrow$} & \textbf{MRR} & \textbf{Hit@1} & \textbf{Hit@5} & \textbf{Hit@10} & \textbf{Prec@1} & \textbf{Prec@5} & \textbf{Prec@10}\\
        \midrule
        TaxoExpan & 1523 & 0.099 & 0.004 & 0.027 & 0.049 & 0.017 & 0.023 & 0.021\\
        ARBORIST & 1142 & 0.133 & 0.008 & 0.044 & 0.075 & 0.037 & 0.038 & 0.033\\
        TMN & \underline{639} & \underline{0.204} & 0.036 & 0.099 & 0.139 & \underline{0.156} & \underline{0.086} & \underline{0.060}\\
        QEN$^\dag$ & 3960 & 0.147 & 0.017 & 0.062 & 0.097 & 0.076 & 0.054 & 0.042\\
        TaxoEnrich* & 5545 & 0.184 & \underline{0.043} & \underline{0.107} & \underline{0.158} & 0.142 &	0.075 &	0.055 \\
        \midrule
        \textsc{TaxBox} & \textbf{596} & \textbf{0.240} & \textbf{0.051} & \textbf{0.139} & \textbf{0.184} & \textbf{0.238} & \textbf{0.131} & \textbf{0.087}\\
        \bottomrule
    \end{tabular}

    \begin{tabular}{c|cccccccc}
        \toprule
         \multirow{2}{*}{\textbf{Method}} & \multicolumn{8}{c}{\textbf{MAG-PSY}} \\
                \cline{2-9}
                & \textbf{MR $\downarrow$} & \textbf{MRR} & \textbf{Hit@1} & \textbf{Hit@5} & \textbf{Hit@10} & \textbf{Prec@1} & \textbf{Prec@5} & \textbf{Prec@10}\\
        \midrule
        TaxoExpan & 728 & 0.253 & 0.015 & 0.092 & 0.163 & 0.031 & 0.038 & 0.033\\
        ARBORIST & 547 & 0.344 & 0.062 & 0.185 & 0.256 & 0.126 & 0.076 & 0.052\\
        TMN & \underline{212} & \underline{0.471} & \underline{0.141} & \underline{0.305} & \underline{0.377} & \underline{0.287} & \underline{0.124} & \underline{0.077}\\
        QEN$^\dag$ & 1778 & 0.293 & 0.103 & 0.150 & 0.206 & 0.103 & 0.059 & 0.042\\
        TaxoEnrich* & 2201 & 0.357 & 0.082 & 0.219 & 0.293 & 0.167 &	0.089	& 0.036 \\
        \midrule
        \textsc{TaxBox} & \textbf{211} & \textbf{0.479} & \textbf{0.145} & \textbf{0.317} & \textbf{0.393} & \textbf{0.328} & \textbf{0.143} & \textbf{0.089}\\
        \bottomrule
    \end{tabular}

    \begin{tabular}{c|cccccccc}
        \toprule
         \multirow{2}{*}{\textbf{Method}} & \multicolumn{8}{c}{\textbf{Wordnet-Verb}} \\
                \cline{2-9}
                & \textbf{MR $\downarrow$} & \textbf{MRR} & \textbf{Hit@1} & \textbf{Hit@5} & \textbf{Hit@10} & \textbf{Prec@1} & \textbf{Prec@5} & \textbf{Prec@10}\\
        \midrule
        TaxoExpan & 1799 & 0.227 & 0.024 & 0.095 & 0.140 & 0.036 & 0.029 & 0.021\\
        ARBORIST & 1637 & 0.206 & 0.016 & 0.073 & 0.116 & 0.024 & 0.022 & 0.018\\
        TMN & \underline{1445} & 0.304 & 0.072 & 0.163 & 0.215 & 0.108 & 0.049 & 0.032\\
        QEN* & 2095 & \textbf{0.331} & \underline{0.074} & \underline{0.178} & \underline{0.233} & \underline{0.113} & \underline{0.054} & \underline{0.036}\\
        TaxoEnrich* & 2873 & 0.320 & 0.069 & 0.168 & 0.229 & 0.106 & 0.052 & 0.035\\
        \midrule
        \textsc{TaxBox} & \textbf{1286} & \underline{0.330} & \textbf{0.105} & \textbf{0.212} & \textbf{0.262} & \textbf{0.179} & \textbf{0.072} & \textbf{0.045}\\
        \bottomrule
    \end{tabular}

    \begin{tabular}{c|cccccccc}
        \toprule
         \multirow{2}{*}{\textbf{Method}} & \multicolumn{8}{c}{\textbf{SemEval-Food}} \\
                \cline{2-9}
                & \textbf{MR $\downarrow$} & \textbf{MRR} & \textbf{Hit@1} & \textbf{Hit@5} & \textbf{Hit@10} & \textbf{Prec@1} & \textbf{Prec@5} & \textbf{Prec@10}\\
        \midrule
        TaxoExpan & 688 & 0.207 & 0.041 & 0.101 & 0.166 & 0.083 & 0.041 & 0.034\\
        ARBORIST & 700 & 0.129 & 0.013 & 0.053 & 0.088 & 0.027 & 0.022 & 0.018\\
        TMN & 559 & 0.211 & 0.037 & 0.113 & 0.160 & 0.074 & 0.046 & 0.032\\
        QEN & 353 & 0.313 & 0.070 & 0.176 & 0.234 & 0.146 & 0.074 & 0.049\\
        TaxoEnrich$^\dag$ & \underline{305} & \underline{0.348} & \underline{0.113} & \underline{0.247} & \underline{0.290} & \underline{0.230} & \underline{0.100} & \underline{0.063} \\
        \midrule
        \textsc{TaxBox} & \textbf{281} & \textbf{0.359} & \textbf{0.132} & \textbf{0.264} & \textbf{0.295} & \textbf{0.318} & \textbf{0.127} & \textbf{0.071}\\
        \bottomrule
    \end{tabular}

    \caption{Overall results on four taxonomy completion datasets. The $\downarrow$ denotes that the lower the metric is the higher performance the model has. Baselines are reported by \citet{zhang2021taxonomy} and \citet{wang2022qen}. * means our reproduction. $\dag$ means our implementation on new datasets. We report the mean results of 5 runs.}
    \label{tab:tab2}
\end{table*}

\noindent \textbf{Ranking Loss}. It's evident that the values of two scorers are numerically unbalanced, namely $S_I \leq S_A$ when considering the same candidate parent. In fact, there is no need for concern, as when a query is inserted into this candidate position, it is implicitly attached as a leaf. Our focus should be on guaranteeing $S_I(X_{pos}) \geq S_A(X_{neg})$ where the subscripts $pos$ and $neg$ indicate positive and negative samples, respectively. Consequently, for $k, k'\in \{I, A\}$, the ranking loss is strategically designed to circumvent this particular case.
\begin{align}
    \begin{split}
        \label{eq:eq13}
    \mathcal{L}_r=\frac{1}{|\mathcal{B}|}\sum_{X_i \in \mathcal{B}} &\text{max}(0,\gamma(X_{pos}, X_{neg})\\
    &+S_k(X_{neg})-S_{k'}(X_{pos}))
    \end{split}
\end{align}
Here, the dynamic margin compels $S_I(X_{pos})$ to be greater than $S_A(X_{neg})$ to a specific extent based on their structural similarity. The final loss combines all of the three losses mentioned above:
\begin{align}
    \label{eq:eq14}
    \mathcal{L} = \mathcal{L}_c +  \mathcal{L}_b + \mathcal{L}_r
\end{align}
\section{Experiments}

\subsection{Experiment Setup}
\noindent \textbf{Datasets}. We assess \textsc{TaxBox}'s performance in taxonomy completion on four real-world datasets: two Microsoft Academic Graph subgraphs, \textit{MAG-CS} and \textit{MAG-PSY}, plus two WordNet subgraphs, \textit{Wordnet-Verb} and \textit{SemEval-Food}. Also, two public datasets from SemEval-16, \textit{Science} and \textit{Environment} are evaluated for taxonomy expansion. Further dataset details are available in Appendix \ref{dataset_details}. Evaluation metrics consist of Mean Rank (MR), Mean Reciprocal Rank (MRR), Wu\&P, Hit@k, and Prec@k, with elaboration in Appendix \ref{eval_metric}.

\noindent \textbf{Compared Methods}. We select three recent SOTA taxonomy completion frameworks, Triplet Matching Network (TMN) \cite{zhang2021taxonomy}, QEN \cite{wang2022qen} and TaxoEnrich\cite{jiang2022taxoenrich}, and two taxonomy expansion frameworks, TaxoExpan \cite{shen2020taxoexpan} and ARBORIST \cite{manzoor2020expanding}, as baselines for the four completion datasets. Additionally, we compare BoxTAXO\cite{boxtaxo} and TaxoExpan demonstrating \textsc{TaxBox}'s superiority in taxonomy expansion. A further explanation is presented in Appendix \ref{compared_methods}.

\noindent \textbf{Implementation Details}. The Adam optimizer was employed with a 0.001 learning rate and the ReduceLROnPlateau scheduler with a 10-epoch patience, training our model across all datasets for 100 epochs. Four attention heads were fixed with 0.1 dropout rate in GAT. The dynamic margin relaxation factor $\alpha$ was 0.5.  The training and prediction smoothness factor $\tau$ were 10 and 20 respectively. Batch and negative sample size were set at 16 and 63, while box dimensions were set at 64 for \textit{SemEval-Food}, 128 for \textit{Wordnet-Verb} and \textit{MAG-CS}, and 160 for \textit{MAG-PSY}. Initial embeddings were the word2vec for the MAG datasets, fasttext for the Wordnet datasets, barring the PLM-based methods, and BERT embedding for two expansion datasets for fair comparison. All the experiments were conducted with one RTX3090.

\subsection{Experimental Results}
Table \ref{tab:tab2} demonstrates the superior performance of \textsc{TaxBox} in taxonomy completion datasets, reflecting average improvements of 6.7\%, 34.9\%, and 51.4\% in MRR, Hit@1, and Prec@1. It outperforms prior SOTA models, such as QEN and TaxoEnrich, which utilize the pre-trained language models (PLM) to enhance the representation. It showcases \textsc{TaxBox}'s performance when handling datasets with varied scales.
\textsc{TaxBox}'s efficacy originates from its box embedding's superior ability to capture asymmetric relationships among concepts and shows a significant improvement over conventional vector representations. PLM-based models like QEN, which lean on rich concepts' descriptions from various internet-based data sources, tend to induce noise, particularly when dealing with larger datasets with obscure, overlapping concepts. Similarly, TaxoEnrich's taxonomy-contextualized embeddings may reveal a variance in distribution between the training and testing phases, chiefly due to the test phase's exclusion of query-related information.

On \textit{MAG-PSY} and \textit{Wordnet-Verb} datasets, \textsc{TaxBox} outperforms in Hit@k and Prec@k metrics but has less exceptional MRR scores. A statistical analysis revealed that in \textit{MAG-CS} and \textit{SemEval-Food} datasets, the ratios of the maximum number of positive candidates in the training set to that in the test set are 2.5 and 1.5, respectively, whereas for \textit{MAG-PSY} and \textit{Wordnet-Verb}, the ratios are 14 and 11. It suggests the need for \textsc{TaxBox} to optimize for all the concept boxes under relatively relaxed conditions to accommodate numerous ground truth positions in the training set. This presents a challenge when identifying test queries with fewer ground truth positions, constricting MRR scores while showing significant improvements in other metrics.


\subsection{Ablation Study}
To assess the efficacy of our proposed learning objectives ($\mathcal{L}_r$, $\mathcal{L}_b$) and graph aggregation module, we performed ablation studies using \textit{SemEval-Food} and \textit{Wordnet-Verb} datasets (Table \ref{tab:tab3}). The model's overall performance deteriorated when any component was removed, more noticeably so with $\mathcal{L}_b$. 
This is due to $\mathcal{L}_b$ explicitly constraining box location and volume, while $\mathcal{L}_r$ primarily balances the gap between scorers, which is implicitly addressed during the optimization process of $\mathcal{L}_c$. Despite that, $\mathcal{L}_r$ still yields a crucial 10\% performance gain. The graph aggregation module demonstrated a significant improvement, underscoring its essential role in enhancing candidate feature enrichment.

\begin{table}[h]
    \centering
    \small
    \begin{tabular}{c|ccc}
        \toprule
         \multirow{2}{*}{\textbf{Method}} & \multicolumn{3}{c}{\textbf{SemEval-Food}} \\
                \cline{2-4}
                 & \textbf{MRR} & \textbf{Hit@1}  & \textbf{Prec@1} \\
        \midrule
        \textsc{TaxBox} w/o $\mathcal{L}_r$ & 0.346 & 0.104 & 0.250 \\
        \textsc{TaxBox} w/o $\mathcal{L}_b$ & 0.304 & 0.084 & 0.202 \\
        \textsc{TaxBox} w/o GAM & 0.347 & 0.112 & 0.270 \\
        \textsc{TaxBox} w/o $\mathcal{L}_b$ \&$\mathcal{L}_r$ & 0.285 & 0.079 & 0.189 \\
        \textsc{TaxBox} & 0.359 & 0.132 & 0.318 \\ 
        \bottomrule
    \end{tabular}
    
    \begin{tabular}{c|ccc}
        \toprule
         \multirow{2}{*}{\textbf{Method}} & \multicolumn{3}{c}{\textbf{Wordnet-Verb}} \\
                \cline{2-4}
                 & \textbf{MRR} & \textbf{Hit@1}  & \textbf{Prec@1} \\
        \midrule
        \textsc{TaxBox} w/o $\mathcal{L}_r$ & 0.316 & 0.097 & 0.165 \\
        \textsc{TaxBox} w/o $\mathcal{L}_b$  & 0.211 & 0.053 & 0.091 \\
        \textsc{TaxBox} w/o GAM & 0.310 & 0.100 & 0.173 \\
        \textsc{TaxBox} w/o $\mathcal{L}_b$\& $\mathcal{L}_r$ & 0.220 & 0.046 & 0.079 \\
        \textsc{TaxBox} & 0.330 & 0.105 & 0.179 \\        
        \bottomrule
    \end{tabular}
    \caption{Ablation study on SemEval-Food and Wordnet-Verb datasets. GAM means graph aggregation module.}
    \label{tab:tab3}
\end{table}


\subsection{How Two Scorers Work for Attachment and Insertion}
Table \ref{tab:tab4} highlights the superior performance of \textsc{TaxBox} over \textit{SemEval-Food} and \textit{Wordnet-Verb} datasets in terms of attachment and insertion, compared to other methods. It excels in all attachment metrics, emphasizing the aptitude of its scorer to utilize box embeddings' spatial aspects, while ignoring child boxes. For insertion, \textsc{TaxBox} outperforms prevailing methods, indicating its scorer's accuracy in identifying optimal candidate positions considering overlap and center similarity. This confirms the effectiveness and necessity of our method, and the insufficiency of pseudo leaf introduction in prior methods.
\begin{table}[h]
    \centering
    \small
    \begin{tabular}{c|cc|cc}
        \toprule
         \multirow{3}{*}{\textbf{Method}} & \multicolumn{4}{c}{\textbf{SemEval-Food}} \\
                \cline{2-5}
                & \multicolumn{2}{c}{Attachment} & \multicolumn{2}{c}{Insertion}\\
                \cline{2-5}
                & \textbf{MRR} & \textbf{Hit@1}  & \textbf{MRR} & \textbf{Hit@1}\\
        \midrule
        TMN  & 0.633 & \underline{0.214}  & 0.069 & 0.000\\
        QEN  & \underline{0.644} & 0.178  & \underline{0.084} & \underline{0.011}\\
        \textsc{TaxBox} & \textbf{0.678} & \textbf{0.288} & \textbf{0.133} & \textbf{0.032}\\
        \bottomrule
    \end{tabular}
    
    \begin{tabular}{c|cc|cc}
        \toprule
         \multirow{3}{*}{\textbf{Method}} & \multicolumn{4}{c}{\textbf{Wordnet-Verb}} \\
                \cline{2-5}
                & \multicolumn{2}{c}{Attachment} & \multicolumn{2}{c}{Insertion}\\
                \cline{2-5}
                 & \textbf{MRR} & \textbf{Hit@1}  & \textbf{MRR} & \textbf{Hit@1}\\
        \midrule
        TMN  & 0.456 & \underline{0.139}  & 0.121 & 0.004\\
        QEN  & \underline{0.466} & 0.125  & \underline{0.160} & \underline{0.007}\\
        \textsc{TaxBox} & \textbf{0.481} & \textbf{0.165}  & \textbf{0.185} & \textbf{0.050}\\
        \bottomrule
    \end{tabular}
    \caption{Performance in attachment and insertion cases.}
    \label{tab:tab4}
\end{table}

\subsection{How \textsc{TaxBox} Solves the Limitation of BoxTAXO}

\begin{table}[h]
    \centering
    \small
    \begin{tabular}{c|ccc}
        \toprule
         \multirow{2}{*}{\textbf{Method}} & \multicolumn{3}{c}{\textbf{Environment}} \\
                \cline{2-4}
                & \textbf{Prec@1} & \textbf{MRR}  & \textbf{Wu\&P}\\
        \midrule
        TaxoExpan  & 11.1 & 32.3  & 54.8\\
        BoxTAXO  & 38.1 & 47.1  & 75.4\\
        \textsc{TaxBox} & \textbf{44.2} & \textbf{55.0} & \textbf{77.8}\\
        \bottomrule
    \end{tabular}
    \begin{tabular}{c|ccc}
        \toprule
         \multirow{2}{*}{\textbf{Method}} & \multicolumn{3}{c}{\textbf{Science}} \\
                \cline{2-4}
                & \textbf{Prec@1} & \textbf{MRR}  & \textbf{Wu\&P}\\
        \midrule
        TaxoExpan  & 27.8 & 44.8  & 57.6\\
        BoxTAXO  & 31.8 & 45.3  & 64.7\\
        \textsc{TaxBox} & \textbf{44.7} & \textbf{54.3} & \textbf{81.3}\\
        \bottomrule
    \end{tabular}
    \caption{The performance of \textsc{TaxBox} on taxonomy expansion datasets. Baselines are reported by \citet{boxtaxo}. *Please note that we have not scaled MRR by 10 and have applied a 100x scale to all results here.}
    \label{tab:tab11}
\end{table}
Table \ref{tab:tab11} reveals that \textsc{TaxBox} surpassed BoxTAXO in all metrics to show the \textsc{TaxBox}'s superiority over BoxTAXO. BoxTAXO's limitations largely stem from its simplification of taxonomies into sheer tree structures, resorting to containment or non-intersection. This approach engenders two primary concerns: 1) Hard boundaries inhibiting multiple parent nodes accommodation, and 2) unreliable inference criteria due to volume containment probability being the chief confidence score. Contrarily, \textsc{TaxBox} mitigates these constraints with its soft margin-based constraints accommodating overlaps, and improves inference criteria with box center-position distance. Consequently, \textsc{TaxBox}'s predictions are more precise, and it capably processes nodes with multiple parents, outperforming BoxTAXO.

\subsection{How Dynamic Margin Affects Box Constraint}
Table \ref{tab:tab12} highlights the dynamic margin's efficiency in box constraint loss, in spite of comparable MRR results. Discrepancies in Hit@1 and Prec@1 across fixed margins accentuate the dynamic margin's superiority in accurately modeling inter-box relationships. While a 0.3 fixed margin in \textit{SemEval-Food} might parallel its performance, determining the optimal margin remains challenging. Notably, the dynamic margin outperforms all fixed margins in \textit{Wordnet-Verb}, further underscoring its adaptability. 
\begin{table}[h]
    \centering
    \small
    \begin{tabular}{c|ccc}
        \toprule
         \multirow{2}{*}{\textbf{Margin}} & \multicolumn{3}{c}{\textbf{SemEval-Food}} \\
                \cline{2-4}
                 & \textbf{MRR} & \textbf{Hit@1}  & \textbf{Prec@1} \\
        \midrule
        0.1 & 0.357 & 0.107 & 0.256 \\
        0.3 & 0.355 & 0.121 & 0.291 \\
        0.5 & 0.352 & 0.104 & 0.250 \\
        dynamic & 0.359 & 0.132 & 0.318 \\ 
        \bottomrule
    \end{tabular}
    \begin{tabular}{c|ccc}
        \toprule
         \multirow{2}{*}{\textbf{Margin}} & \multicolumn{3}{c}{\textbf{Wordnet-Verb}} \\
                \cline{2-4}
                 & \textbf{MRR} & \textbf{Hit@1}  & \textbf{Prec@1} \\
        \midrule
        0.1 & 0.318 & 0.096 & 0.164 \\
        0.3 & 0.328 & 0.092 & 0.157 \\
        0.5 & 0.322 & 0.090 & 0.154 \\
        dynamic & 0.330 & 0.105 & 0.179 \\ 
        \bottomrule
    \end{tabular}
    \caption{The results of different margins in the box constrain loss on two datasets.}
    \label{tab:tab12}
\end{table}

\subsection{How to set up \textsc{TaxBox}}
We discuss our choice for box dimensionality and the number of negative samples in Appendix \ref{hyper}.

\section{Conclusion}
In this study, we present \textsc{TaxBox}, a novel framework for taxonomy completion using box embeddings. Incorporating restricted box constraint loss, dynamic ranking loss, and two probabilistic scorers for attachment and insertion, \textsc{TaxBox} employs a structurally enhanced box decoder, mitigating the need for pseudo leaves. Experiments on six real-world datasets demonstrate its effectiveness and performance. Future research could refine scorers without numerical imbalance and explore post-processing measures like reranking with LLM.

\clearpage

\section*{Limitations}
The primary limitations of our proposed methods are as follows: (1) The numerical imbalance between the two scorers. Although we attempt to alleviate this issue by introducing a dynamic ranking loss, it remains an imperfect solution. Results shown in Table \ref{tab:tab4} indicate that tackling the insertion case in real-world practice is still challenging, despite \textsc{TaxBox} achieving significant improvements compared to previous SOTA. A more practical scorer should be developed to address this. (2) In real-world applications, the quality of the initial embedding influences \textsc{TaxBox}'s performance to some extent. Even when we opt for a well-pretrained language model for encoding, the concept name and description have a considerable impact. Thus, a more adaptive training strategy is needed. For example, we could employ data augmentation techniques to generate multiple texts representing the same meaning and use a PLM to obtain an embedding set pointing to a specific concept. During training, we can then retrieve different embeddings to fit the network, consequently enhancing its generalization capabilities.

\section*{Acknowledgments}

This work is supported by the National Natural Science Foundation of China (No. 62376245), the Key Research and Development Program of Zhejiang Province, China (No. 2024C01034), the project of the Donghai Laboratory (Grant no. DH-2022ZY0013),  National Key Research and Development Project of China (No. 2018AAA0101900), and MOE Engineering Research Center of Digital Library.

\bibliography{custom}

\begin{thebibliography}{39}
\expandafter\ifx\csname natexlab\endcsname\relax\def\natexlab#1{#1}\fi

\bibitem[{Arous et~al.(2023)Arous, Dolamic, and Cudr{\'e}-Mauroux}]{arous2023taxocomplete}
Ines Arous, Ljiljana Dolamic, and Philippe Cudr{\'e}-Mauroux. 2023.
\newblock Taxocomplete: Self-supervised taxonomy completion leveraging position-enhanced semantic matching.
\newblock In \emph{Proceedings of the ACM Web Conference 2023}, pages 2509--2518.

\bibitem[{Bordea et~al.(2015)Bordea, Buitelaar, Faralli, and Navigli}]{bordea2015semeval17}
Georgeta Bordea, Paul Buitelaar, Stefano Faralli, and Roberto Navigli. 2015.
\newblock Semeval-2015 task 17: Taxonomy extraction evaluation (texeval).
\newblock In \emph{Proceedings of the 9th International Workshop on Semantic Evaluation (SemEval 2015)}, pages 902--910.

\bibitem[{Chatterjee and Das(2022)}]{chatterjee2022analysing}
Swarnali Chatterjee and Rajesh Das. 2022.
\newblock Analysing and examining taxonomy and folksonomy terms in the hybrid subject device using machine learning techniques.
\newblock \emph{DESIDOC Journal of Library \& Information Technology}, 42(3):154.

\bibitem[{Chheda et~al.(2021)Chheda, Goyal, Tran, Patel, Boratko, Dasgupta, and McCallum}]{chheda2021box}
Tejas Chheda, Purujit Goyal, Trang Tran, Dhruvesh Patel, Michael Boratko, Shib~Sankar Dasgupta, and Andrew McCallum. 2021.
\newblock Box embeddings: An open-source library for representation learning using geometric structures.
\newblock \emph{arXiv preprint arXiv:2109.04997}.

\bibitem[{Chuang and Chien(2003)}]{chuang2003automatic}
Shui-Lung Chuang and Lee-Feng Chien. 2003.
\newblock Automatic query taxonomy generation for information retrieval applications.
\newblock \emph{Online Information Review}.

\bibitem[{Dasgupta et~al.(2022)Dasgupta, Boratko, Mishra, Atmakuri, Patel, Li, and McCallum}]{dasgupta2022word2box}
Shib Dasgupta, Michael Boratko, Siddhartha Mishra, Shriya Atmakuri, Dhruvesh Patel, Xiang Li, and Andrew McCallum. 2022.
\newblock Word2box: Capturing set-theoretic semantics of words using box embeddings.
\newblock In \emph{Proceedings of the 60th Annual Meeting of the Association for Computational Linguistics (Volume 1: Long Papers)}, pages 2263--2276.

\bibitem[{Devlin et~al.(2018)Devlin, Chang, Lee, and Toutanova}]{devlin2018bert}
Jacob Devlin, Ming-Wei Chang, Kenton Lee, and Kristina Toutanova. 2018.
\newblock Bert: Pre-training of deep bidirectional transformers for language understanding.
\newblock \emph{arXiv preprint arXiv:1810.04805}.

\bibitem[{Huang et~al.(2019)Huang, Ren, Zhao, He, Wen, and Dong}]{huang2019taxo}
Jin Huang, Zhaochun Ren, Wayne~Xin Zhao, Gaole He, Ji-Rong Wen, and Daxiang Dong. 2019.
\newblock \href {https://doi.org/10.1145/3289600.3290972} {Taxonomy-aware multi-hop reasoning networks for sequential recommendation}.
\newblock New York, NY, USA. Association for Computing Machinery.

\bibitem[{Hwang et~al.(2022)Hwang, Lee, Yang, Patel, Zhang, and McCallum}]{hwang2022event}
EunJeong Hwang, Jay-Yoon Lee, Tianyi Yang, Dhruvesh Patel, Dongxu Zhang, and Andrew McCallum. 2022.
\newblock Event-event relation extraction using probabilistic box embedding.
\newblock In \emph{Proceedings of the 60th Annual Meeting of the Association for Computational Linguistics (Volume 2: Short Papers)}, pages 235--244.

\bibitem[{Jiang et~al.(2022)Jiang, Song, Zhang, and Han}]{jiang2022taxoenrich}
Minhao Jiang, Xiangchen Song, Jieyu Zhang, and Jiawei Han. 2022.
\newblock Taxoenrich: Self-supervised taxonomy completion via structure-semantic representations.
\newblock In \emph{Proceedings of the ACM Web Conference 2022}, pages 925--934.

\bibitem[{Jiang et~al.(2023)Jiang, Yao, Wang, and Sun}]{boxtaxo}
Song Jiang, Qiyue Yao, Qifan Wang, and Yizhou Sun. 2023.
\newblock A single vector is not enough: Taxonomy expansion via box embeddings.
\newblock In \emph{Proceedings of the ACM Web Conference 2023}, pages 2467--2476.

\bibitem[{Jurgens and Pilehvar(2016)}]{jurgens2016semeval14}
David Jurgens and Mohammad~Taher Pilehvar. 2016.
\newblock Semeval-2016 task 14: Semantic taxonomy enrichment.
\newblock In \emph{Proceedings of the 10th international workshop on semantic evaluation (SemEval-2016)}, pages 1092--1102.

\bibitem[{Kejriwal et~al.(2022)Kejriwal, Selvam, Ni, and Torzec}]{kejriwal2022local}
Mayank Kejriwal, Ravi~Kiran Selvam, Chien-Chun Ni, and Nicolas Torzec. 2022.
\newblock Local taxonomy construction: An information retrieval approach using representation learning.
\newblock In \emph{Social Media Analysis for Event Detection}, pages 133--161. Springer.

\bibitem[{Kerschberg et~al.(2001)Kerschberg, Kim, and Scime}]{Kerschberg2001tax}
L.~Kerschberg, Wooju Kim, and A.~Scime. 2001.
\newblock \href {https://doi.org/10.1109/WISE.2001.996465} {A semantic taxonomy-based personalizable meta-search agent}.
\newblock In \emph{Proceedings of the Second International Conference on Web Information Systems Engineering}, volume~1, pages 41--50 vol.1.

\bibitem[{Liu et~al.(2021)Liu, Xu, Wen, Jiang, Wu, and Yuan}]{liu-etal-2021-temp}
Zichen Liu, Hongyuan Xu, Yanlong Wen, Ning Jiang, HaiYing Wu, and Xiaojie Yuan. 2021.
\newblock \href {https://doi.org/10.18653/v1/2021.emnlp-main.313} {{TEMP}: Taxonomy expansion with dynamic margin loss through taxonomy-paths}.
\newblock In \emph{Proceedings of the 2021 Conference on Empirical Methods in Natural Language Processing}, pages 3854--3863, Online and Punta Cana, Dominican Republic. Association for Computational Linguistics.

\bibitem[{Ma et~al.(2021)Ma, Chen, Wu, and Peng}]{ma2021hyperexpan}
Mingyu~Derek Ma, Muhao Chen, Te-Lin Wu, and Nanyun Peng. 2021.
\newblock Hyperexpan: Taxonomy expansion with hyperbolic representation learning.
\newblock In \emph{Findings of the Association for Computational Linguistics: EMNLP 2021}, pages 4182--4194.

\bibitem[{Manzoor et~al.(2020)Manzoor, Li, Shrouty, and Leskovec}]{manzoor2020expanding}
Emaad Manzoor, Rui Li, Dhananjay Shrouty, and Jure Leskovec. 2020.
\newblock Expanding taxonomies with implicit edge semantics.
\newblock In \emph{Proceedings of The Web Conference 2020}, pages 2044--2054.

\bibitem[{Messner et~al.(2022)Messner, Abboud, and Ceylan}]{messner2022temporal}
Johannes Messner, Ralph Abboud, and Ismail~Ilkan Ceylan. 2022.
\newblock Temporal knowledge graph completion using box embeddings.
\newblock In \emph{Proceedings of the AAAI Conference on Artificial Intelligence}, volume~36, pages 7779--7787.

\bibitem[{Miller(1995)}]{miller1995wordnet}
George~A Miller. 1995.
\newblock Wordnet: a lexical database for english.
\newblock \emph{Communications of the ACM}, 38(11):39--41.

\bibitem[{Onoe et~al.(2021)Onoe, Boratko, McCallum, and Durrett}]{onoe2021modeling}
Yasumasa Onoe, Michael Boratko, Andrew McCallum, and Greg Durrett. 2021.
\newblock Modeling fine-grained entity types with box embeddings.
\newblock In \emph{Proceedings of the 59th Annual Meeting of the Association for Computational Linguistics and the 11th International Joint Conference on Natural Language Processing (Volume 1: Long Papers)}, pages 2051--2064.

\bibitem[{Patel et~al.(2021)Patel, Dangati, Lee, Boratko, and McCallum}]{patel2021modeling}
Dhruvesh Patel, Pavitra Dangati, Jay-Yoon Lee, Michael Boratko, and Andrew McCallum. 2021.
\newblock Modeling label space interactions in multi-label classification using box embeddings.
\newblock In \emph{International Conference on Learning Representations}.

\bibitem[{Phukon et~al.(2022)Phukon, Mitra, Sanasam, and Sarmah}]{phukon2022team}
Bornali Phukon, Anasua Mitra, Ranbir Sanasam, and Priyankoo Sarmah. 2022.
\newblock Team: A multitask learning based taxonomy expansion approach for attach and merge.
\newblock In \emph{Findings of the Association for Computational Linguistics: NAACL 2022}, pages 366--378.

\bibitem[{Rau et~al.(2020)Rau, Garcia-Hernando, Stoyanov, Brostow, and Turmukhambetov}]{rau2020predicting}
Anita Rau, Guillermo Garcia-Hernando, Danail Stoyanov, Gabriel~J Brostow, and Daniyar Turmukhambetov. 2020.
\newblock Predicting visual overlap of images through interpretable non-metric box embeddings.
\newblock In \emph{European Conference on Computer Vision}, pages 629--646. Springer.

\bibitem[{Sanh et~al.(2019)Sanh, Debut, Chaumond, and Wolf}]{sanh2019distilbert}
Victor Sanh, Lysandre Debut, Julien Chaumond, and Thomas Wolf. 2019.
\newblock Distilbert, a distilled version of bert: smaller, faster, cheaper and lighter.
\newblock \emph{arXiv preprint arXiv:1910.01108}.

\bibitem[{Shen et~al.(2020)Shen, Shen, Xiong, Wang, Wang, and Han}]{shen2020taxoexpan}
Jiaming Shen, Zhihong Shen, Chenyan Xiong, Chi Wang, Kuansan Wang, and Jiawei Han. 2020.
\newblock \href {https://doi.org/10.1145/3366423.3380132} {Taxoexpan: Self-supervised taxonomy expansion with position-enhanced graph neural network}.
\newblock In \emph{Proceedings of The Web Conference 2020}, WWW '20, page 486–497, New York, NY, USA. Association for Computing Machinery.

\bibitem[{Shen et~al.(2018)Shen, Wu, Lei, Zhang, Ren, Vanni, Sadler, and Han}]{shen2018hiexpan}
Jiaming Shen, Zeqiu Wu, Dongming Lei, Chao Zhang, Xiang Ren, Michelle~T. Vanni, Brian~M. Sadler, and Jiawei Han. 2018.
\newblock \href {https://doi.org/10.1145/3219819.3220115} {Hiexpan: Task-guided taxonomy construction by hierarchical tree expansion}.
\newblock In \emph{Proceedings of the 24th ACM SIGKDD International Conference on Knowledge Discovery Data Mining}, KDD '18, page 2180–2189, New York, NY, USA. Association for Computing Machinery.

\bibitem[{Sinha et~al.(2015)Sinha, Shen, Song, Ma, Eide, Hsu, and Wang}]{sinha2015overview}
Arnab Sinha, Zhihong Shen, Yang Song, Hao Ma, Darrin Eide, Bo-June Hsu, and Kuansan Wang. 2015.
\newblock An overview of microsoft academic service (mas) and applications.
\newblock In \emph{Proceedings of the 24th international conference on world wide web}, pages 243--246.

\bibitem[{Srivastava et~al.(2015)Srivastava, Greff, and Schmidhuber}]{srivastava2015highway}
Rupesh~Kumar Srivastava, Klaus Greff, and J{\"u}rgen Schmidhuber. 2015.
\newblock Highway networks.
\newblock \emph{arXiv preprint arXiv:1505.00387}.

\bibitem[{Staudemeyer and Morris(2019)}]{staudemeyer2019lstm}
Ralf~C Staudemeyer and Eric~Rothstein Morris. 2019.
\newblock Understanding lstm--a tutorial into long short-term memory recurrent neural networks.
\newblock \emph{arXiv preprint arXiv:1909.09586}.

\bibitem[{Suchanek et~al.(2007)Suchanek, Kasneci, and Weikum}]{suchanek2007yago}
Fabian~M Suchanek, Gjergji Kasneci, and Gerhard Weikum. 2007.
\newblock Yago: a core of semantic knowledge.
\newblock In \emph{Proceedings of the 16th international conference on World Wide Web}, pages 697--706.

\bibitem[{Veličković et~al.(2018)Veličković, Cucurull, Casanova, Romero, Liò, and Bengio}]{gat}
Petar Veličković, Guillem Cucurull, Arantxa Casanova, Adriana Romero, Pietro Liò, and Yoshua Bengio. 2018.
\newblock \href {https://openreview.net/forum?id=rJXMpikCZ} {Graph attention networks}.
\newblock In \emph{International Conference on Learning Representations}.

\bibitem[{Vilnis et~al.(2018)Vilnis, Li, Murty, and McCallum}]{vilnis2018probabilistic}
Luke Vilnis, Xiang Li, Shikhar Murty, and Andrew McCallum. 2018.
\newblock Probabilistic embedding of knowledge graphs with box lattice measures.
\newblock \emph{arXiv preprint arXiv:1805.06627}.

\bibitem[{Wang et~al.(2022)Wang, Zhao, Zheng, and Liu}]{wang2022qen}
Suyuchen Wang, Ruihui Zhao, Yefeng Zheng, and Bang Liu. 2022.
\newblock \href {https://doi.org/10.1145/3485447.3511943} {Qen: Applicable taxonomy completion via evaluating full taxonomic relations}.
\newblock In \emph{Proceedings of the ACM Web Conference 2022}, WWW '22, page 1008–1017, New York, NY, USA. Association for Computing Machinery.

\bibitem[{Wu and Palmer(1994)}]{wu1994verb}
Zhibiao Wu and Martha Palmer. 1994.
\newblock Verb semantics and lexical selection.
\newblock In \emph{32nd Annual Meeting of the Association for Computational Linguistics}, pages 133--138.

\bibitem[{Yang et~al.(2017)Yang, Zou, Wang, Yan, and Wen}]{yang2017ea}
Shuo Yang, Lei Zou, Zhongyuan Wang, Jun Yan, and Ji-Rong Wen. 2017.
\newblock \href {https://doi.org/10.1609/aaai.v31i1.10956} {Efficiently answering technical questions — a knowledge graph approach}.
\newblock \emph{Proceedings of the AAAI Conference on Artificial Intelligence}, 31(1).

\bibitem[{Yu et~al.(2021)Yu, Wu, Deng, Zeng, Mahindru, Guven, and Jiang}]{yu-etal-2021-technical}
Wenhao Yu, Lingfei Wu, Yu~Deng, Qingkai Zeng, Ruchi Mahindru, Sinem Guven, and Meng Jiang. 2021.
\newblock \href {https://doi.org/10.18653/v1/2021.naacl-industry.23} {Technical question answering across tasks and domains}.
\newblock In \emph{Proceedings of the 2021 Conference of the North American Chapter of the Association for Computational Linguistics: Human Language Technologies: Industry Papers}, pages 178--186, Online. Association for Computational Linguistics.

\bibitem[{Yu et~al.(2020)Yu, Li, Shen, Feng, Sun, and Zhang}]{yu2020steam}
Yue Yu, Yinghao Li, Jiaming Shen, Hao Feng, Jimeng Sun, and Chao Zhang. 2020.
\newblock Steam: Self-supervised taxonomy expansion with mini-paths.
\newblock In \emph{Proceedings of the 26th ACM SIGKDD International Conference on Knowledge Discovery \& Data Mining}, pages 1026--1035.

\bibitem[{Zeng et~al.(2021)Zeng, Lin, Yu, Cleland-Huang, and Jiang}]{zeng2021taxogen}
Qingkai Zeng, Jinfeng Lin, Wenhao Yu, Jane Cleland-Huang, and Meng Jiang. 2021.
\newblock \href {https://doi.org/10.1145/3447548.3467308} {Enhancing taxonomy completion with concept generation via fusing relational representations}.
\newblock In \emph{Proceedings of the 27th ACM SIGKDD Conference on Knowledge Discovery Data Mining}, KDD '21, page 2104–2113, New York, NY, USA. Association for Computing Machinery.

\bibitem[{Zhang et~al.(2021)Zhang, Song, Zeng, Chen, Shen, Mao, and Li}]{zhang2021taxonomy}
Jieyu Zhang, Xiangchen Song, Ying Zeng, Jiaze Chen, Jiaming Shen, Yuning Mao, and Lei Li. 2021.
\newblock Taxonomy completion via triplet matching network.
\newblock In \emph{Proceedings of the AAAI Conference on Artificial Intelligence}, volume~35, pages 4662--4670.

\end{thebibliography}

\appendix
\section{Dataset}
\label{dataset_details}
We choose six real-world English datasets in different domains, four for taxonomy completion and two for taxonomy expansion. The statistical information about six datasets is shown in table \ref{tab:tab1}.
\begin{itemize}
    \item \textbf{Microsoft Academic Graph (MAG)}~\cite{sinha2015overview} is a large, multi-disciplinary graph. The data in MAG includes information from a wide range of academic disciplines and includes more than 660 thousand scientific concepts and more than 700 thousand taxonomic relations. Following \citet{zhang2021taxonomy}, we use subgraphs related to the computer science (\textbf{MAG-CS}) and psychology(\textbf{MAG-PSY}) domains. The initial embedding is a 250-dimension word2vec embedding trained by \citet{zhang2021taxonomy}.
    \item \textbf{Wordnet}~\cite{miller1995wordnet} is a large lexical database of English. Following ~\cite{wang2022qen} and ~\cite{zhang2021taxonomy}, we choose \textbf{Wordnet-Verb}~\cite{jurgens2016semeval14} and \textbf{SemEval-Food}~\cite{bordea2015semeval17} which are extracted from wordnet. We employ 300-dimension fasttext embedding as our initial features following \citet{zhang2021taxonomy}.
    \item  \textbf{SemEval-16} we use two public datasets released from SemEval-16 task. Specifically, they are small-scaled taxonomy in the domains of \textbf{Environment} and general \textbf{Science}. And their initial embeddings are produced by a pre-trained bert\cite{devlin2018bert}.
\end{itemize}
For \textit{MAG-CS}, \textit{MAG-PSY} and \textit{Wordnet-Verb}, we randomly select 1,000 nodes for testing and 1,000 nodes for validation in each dataset, following the approach of \citet{zhang2021taxonomy}. For \textit{SemEval-Food}, we sample 10\% of all the nodes for testing and another 10\% for validation as done by \citet{wang2022qen}. For \textit{Environment} and \textit{Science}, we adopt the same protocol by \citet{boxtaxo}. Subsequently, we reconstruct the seed taxonomy using the remaining nodes and add edges between the parent and child nodes of the test and validation sets to restore the fragmented taxonomy resulting from the dataset split.

\begin{table} 
    \centering
  
  \begin{tabular}{cccc}
    \toprule
    Dataset&|$\mathcal{N}$|&|$\mathcal{E}$|&|C|\\
    \midrule
    MAG-CS & 24,754 & 42,329 & 153,726\\
    MAG-PSY & 23,187 & 30,041 & 101,077\\
    Wordnet-Verb & 13,936 & 13,408 & 51,159\\
    SemEval-Food & 1,486 & 1,533 & 6,122\\
    Science & 344 & 354 & 344\\
    Environment &  209 & 209 & 209\\
  \bottomrule
\end{tabular}
\caption{The statistics of six datasets. |$\mathcal{N}$|, |$\mathcal{E}$|, |C| are the number of nodes, edges, and candidate positions, respectively.}
\label{tab:tab1}
\end{table}

\section{Evaluation Metric}
\label{eval_metric}
All the methods as well as our model are ranking-based ones, so we use the ranking-based metric to evaluate performance. Supposing $rank(c_i)$ denotes the predicted rank of ground truth position given a query concept $c_i \in C$:

\begin{itemize}
    \item \textbf{Mean Rank (MR)} mainly measures the average tail ranking level and we first calculate the average rank positions of each query and then average all the queries:
    \begin{align}
        \text{MR}=\frac{1}{|C|}\sum_{i=1}^{|C|}(\frac{1}{M_i}\sum_{j=1}^{M_i}\text{rank}(c_i^j))
    \end{align}
    where $M_i$ denotes the total number of ground truth positions of a query $c_i$ and $c_i^j$ denotes the $j$th prediction of $c_i$.
    \item \textbf{Mean Reciprocal Rank (MRR)} mainly measures the average head ranking level. Its form is similar to MR except that we get the reciprocal number of the ranks. Here we scale the reciprocal rank by 10 to amplify the difference.
    \begin{align}
        \text{RR}=\frac{1}{M_i}\sum_{j=1}^{M_i}&\frac{1}{\text{max}(1, \text{rank}(c_i^j) / 10)}\\
        \text{MRR}=&\frac{1}{|C|}\sum_{i=1}^{|C|}\text{RR}
    \end{align}
    \item \textbf{Hit@k} measures the recall of a model which averages the true rank positions for all queries in top $k$:
    \begin{align}
    \text{Hit@k}=\frac{\sum_{i=1}^{|C|}\sum_{j=1}^{M_i}\mathbbm{1}(\text{rank}(c_i^j) \leq \text{k})}{\sum_{i=1}^{|C|} M_i}
    \end{align}
    \item \textbf{Prec@k} measures the precision of the results and it sums the true rank positions of all queries in top $k$, divided by k times the total number of queries:
    \begin{align}
    \text{Prec@k}=\frac{\sum_{i=1}^{|C|}\sum_{j=1}^{M_i}\mathbbm{1}(\text{rank}(c_i^j) \leq \text{k})}{\text{k}*|C|}
    \end{align}
    \item \textbf{Wu\&P}\cite{wu1994verb} measures the structural similarity: 
    \begin{align}
    \text{Wu$\&$P}= \frac{1}{|C|}\sum_{i=1}^{|C|} \frac{2\times \text{depth}(\text{LCA}(a_i,b_i))}{\text{depth}(a_i)+\text{depth}(b_i)}
    \end{align}
    where $a_i$ and $b_i$ are the predicted top-1 result and the truth potision in taxonomy.
\end{itemize}

\section{Compared Methods}
\label{compared_methods}
Here are the details of compared models:
\begin{itemize}
    \item \textbf{TaxoExpan}~\cite{shen2020taxoexpan}: a state-of-the-art method in taxonomy expansion that utilizes a graph neural network to incorporate structural information.
    \item \textbf{ARBORIST}~\cite{manzoor2020expanding}: a state-of-the-art framework for taxonomy expansion and it leverages heterogeneous edge semantics with a dynamic margin loss.
    \item \textbf{BoxTAXO}~\cite{boxtaxo}: a state-of-the-art method using the property of conditional probability of box embedding for taxonomy expansion.
    \item \textbf{TMN}~\cite{zhang2021taxonomy}: a state-of-the-art method for taxonomy completion that employs the channel-wise gate mechanism and auxiliary learning with multiple NTNs to evaluate partially positive candidate pairs beside positive pairs.
    \item \textbf{QEN}~\cite{wang2022qen}: a state-of-the-art model for taxonomy completion which utilizes a pre-trained language model to enhance the initial embedding with semantically rich term representation and enhance the performance with a sibling detector.
    \item \textbf{TaxoEnrich}~\cite{jiang2022taxoenrich}: a state-of-the-art model for taxonomy completion that leverages Taxonomy-Contextualized Embeddings and sibling matching modules.
\end{itemize}


\begin{figure*}[htbp]
    \centering
    \includegraphics[width=\linewidth]{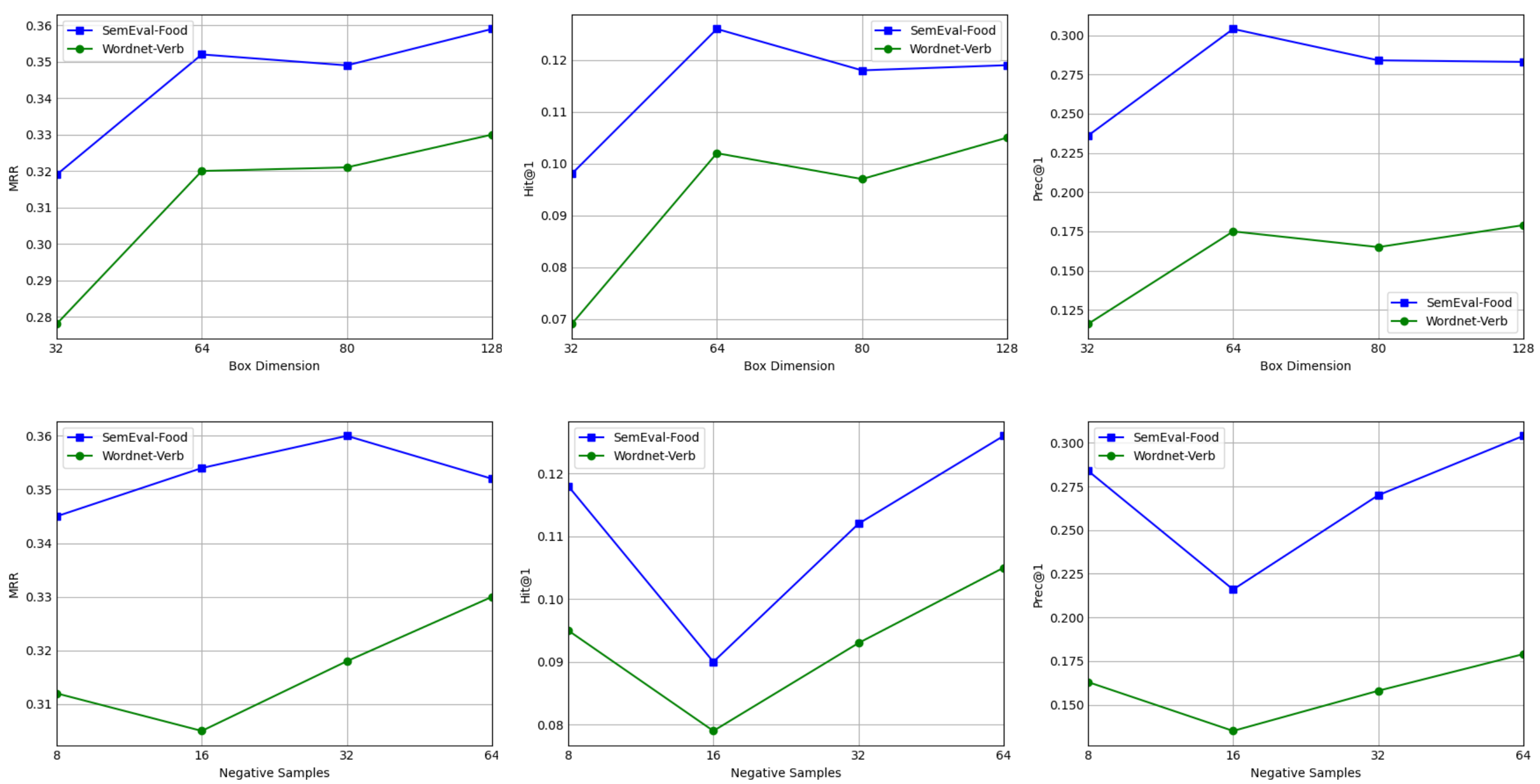}
    \captionof{figure}{The effect of box dimensionality and the number of negative samples over three datasets.}
   \label{fig:fig3}
\end{figure*}

\section{the Effect of Box Dimensionality and Negative Samples}
\label{hyper}
We are also interested in how the box dimensionality and the number of negative samples affect the performance. Figure \ref{fig:fig3} shows the results of MRR, Hit@1 and Prec@1 when changing the box dimensionality from \{ 32, 64, 80, 128 \} and the total number of samples from \{ 8, 16, 32, 64 \}(where negative samples are \{ 7, 15, 31, 63 \}) over two datasets.

Notably, it can be observed that for small datasets \textit{SemEval-Food}, a dimension of 64 serves as a turning point. Dimensions below 64 exhibit a significant decline in overall performance. On the other hand, dimensions exceeding 64 reach a plateau, indicating that 64 is an appropriate dimension. Furthermore, increasing the dimension beyond 64 does not yield further performance improvements; instead, it leads to a decrease. This can be attributed to the fact that a dimension of 64 already satisfies the spatial constraints for all boxes in such a scale dataset. Larger dimensions introduce redundancy, thereby increasing the optimization difficulty. However, for \textit{Wordnet-Verb}, it is worth noting that there is still some performance improvement observed after surpassing 64 dimensions. This discrepancy can be attributed to the larger dataset size and the initial quality of embeddings, which require more dimensions to effectively accommodate the information.

Regarding the setting of negative sample quantities, a general observation can be made that larger numbers of negative samples result in better performance on both datasets. However, it is crucial to acknowledge that an increased number of negative samples reduces the attention given to positive samples during the optimization process of the classification loss. Consequently, it becomes necessary to elevate the weight assigned to positive samples in calculations. Therefore, the steep decrease observed at the position of 16 is a consequence of equal weighting given to positive and negative samples in the experiment, while higher negative sample counts were assigned higher weights. This emphasizes the significance of appropriately adjusting the weight allocation to balance the impact of positive and negative samples during training.

\end{document}